Original Paper

# Explainable Artificial Intelligence Recommendation System by Leveraging the Semantics of Adverse Childhood Experiences: Proof-of-Concept Prototype Development

Nariman Ammar, PhD; Arash Shaban-Nejad, MSc, MPH, PhD

University of Tennessee Health Science Center - Oak Ridge National Laboratory, Center for Biomedical Informatics, Department of Pediatrics, College of Medicine, Memphis, TN, United States

**Corresponding Author:**
Arash Shaban-Nejad, MSc, MPH, PhD
University of Tennessee Health Science Center - Oak Ridge National Laboratory, Center for Biomedical Informatics
Department of Pediatrics, College of Medicine
Memphis, TN
United States
Phone: 1 901 287 583
Email: ashabann@uthsc.edu

## Abstract

**Background:** The study of adverse childhood experiences and their consequences has emerged over the past 20 years. Although the conclusions from these studies are available, the same is not true of the data. Accordingly, it is a complex problem to build a training set and develop machine-learning models from these studies. Classic machine learning and artificial intelligence techniques cannot provide a full scientific understanding of the inner workings of the underlying models. This raises credibility issues due to the lack of transparency and generalizability. Explainable artificial intelligence is an emerging approach for promoting credibility, accountability, and trust in mission-critical areas such as medicine by combining machine-learning approaches with explanatory techniques that explicitly show what the decision criteria are and why (or how) they have been established. Hence, thinking about how machine learning could benefit from knowledge graphs that combine "common sense" knowledge as well as semantic reasoning and causality models is a potential solution to this problem.

**Objective:** In this study, we aimed to leverage explainable artificial intelligence, and propose a proof-of-concept prototype for a knowledge-driven evidence-based recommendation system to improve mental health surveillance.

**Methods:** We used concepts from an ontology that we have developed to build and train a question-answering agent using the Google DialogFlow engine. In addition to the question-answering agent, the initial prototype includes knowledge graph generation and recommendation components that leverage third-party graph technology.

**Results:** To showcase the framework functionalities, we here present a prototype design and demonstrate the main features through four use case scenarios motivated by an initiative currently implemented at a children's hospital in Memphis, Tennessee. Ongoing development of the prototype requires implementing an optimization algorithm of the recommendations, incorporating a privacy layer through a personal health library, and conducting a clinical trial to assess both usability and usefulness of the implementation.

**Conclusions:** This semantic-driven explainable artificial intelligence prototype can enhance health care practitioners' ability to provide explanations for the decisions they make.

(*JMIR Med Inform 2020;8(11):e18752*)   doi: 10.2196/18752

## Introduction

**Background**

The concept of adverse childhood experiences (ACEs) has been recognized for quite some time but was first formally studied in the CDC-Kaiser landmark study [1], which uncovered the strong connection between ACEs and the development of risk factors for different negative health outcomes that threaten the well-being of populations throughout their life course. Social determinants of health (SDoH) are measurable indicators of social conditions in which a patient is embedded. Individuals who experience a more negative burden of these factors within their neighborhood are at higher risk of negative health outcomes [2-4]. There is an entire body of research focused on studying the links between ACEs and SDoH and health outcomes, but few intelligent tools are available to assist in the real-time screening of patients and to assess the connection between ACEs and SDoH, which could help to guide patients and families to available resources (eg, health care providers, government, and nongovernment agencies). Other recent works have focused on developing question-answering (QA) systems for training nursing practitioners on how to answer patient inquiries [5].

Recommendation systems and digital assistants often require machine learning (ML), artificial intelligence (AI), and natural language processing capabilities to effectively connect and harvest the vast amounts of generated data. They also need to store, retrieve, and learn from past interactions and experiences with users. Traditionally, recommendation systems have relied on classic ML techniques that often cannot provide a full scientific understanding of the inner workings of the underlying models. For AI to mimic human intelligence, it needs to incorporate one of the most important classes of information people use to make predictions and decisions: context. Although a standard AI algorithm can learn useful rules from a training set, it also tends to learn other unnecessary, nongeneralizable rules, which may lead to a lack of consistency, transparency, and generalizability. Explainable AI is an emerging approach for promoting credibility, accountability, and trust in mission-critical areas such as medicine by combining ML techniques with explanatory techniques that explicitly show why a recommendation is made. One way to achieve this is by considering formal ontologies as an integral part of the learning process, providing the necessary contextual knowledge about a phenomenon. AI and ML become more trustworthy when underpinned by contextual information provided by ontological platforms. Ontologies can be represented through a graph structure. When computations are performed over graph-structured data, they can make generalizations related to structure rather than data since they support relational reasoning and combinatorial generalization [6]. Moreover, when AI apps are based on contextually aware and dynamic backends, they become easier to train with minimal maintenance. A knowledge graph [7] can serve as a dynamic backend that stores data in a certain domain as entities and relationships using a graph model, which abides by an ontology. Several studies have incorporated graph technologies into ML models applied to biomedical informatics problems [8-11].

We here propose the Semantic Platform for Adverse Childhood Experiences Surveillance (SPACES), an intelligent recommendation system that employs ML techniques to help in screening patients and allocating or discovering relevant resources. The novelty in the approach lies in its ability to use the contextual knowledge collected about the user, and infer new knowledge to support subsequent QA and resource allocations during intake assessment sessions in (near) real time. Moreover, our proposed system intends to build rapport with patients by generating personalized questions during interviews while minimizing the amount of information that needs to be collected directly from the patient.

In our previous work [12-14], we developed the Adverse Childhood Experiences Ontology (ACESO) that captures knowledge on ACEs, SDoH, health outcomes, and interventions. Both expressive and light versions of the ontology can be freely downloaded via BioPortal [15]. The ontology defines concept and property hierarchies, and encodes causal epidemiological knowledge as axioms. We also developed a repository termed Urban Population Health Observatory (UPHO), which provides metrics based on several socioeconomic and environmental data at the neighborhood level that can be linked to health outcomes. Figure 1 demonstrates the multiple layers, from data to the app, that are required for building a knowledge-based explainable and interpretable model.

**Figure 1.** Multilayer representation of a knowledge-based explainable/interpretable model.

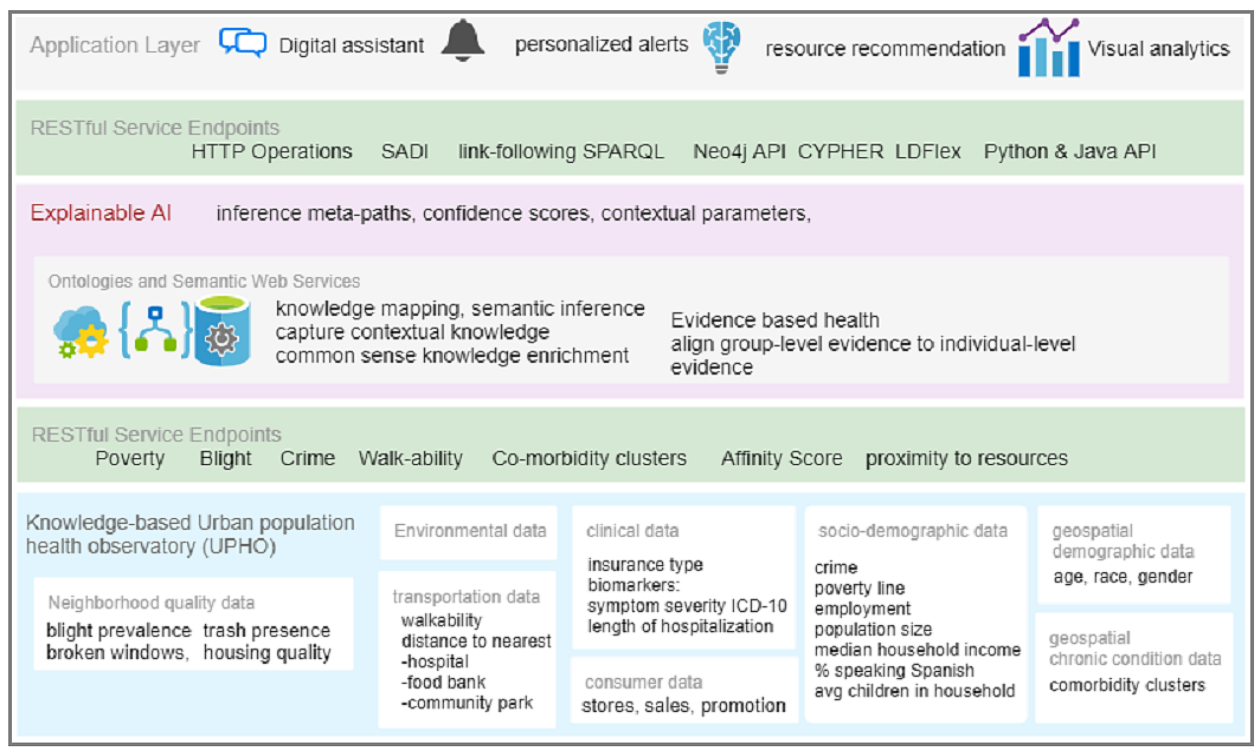

### ACESO

There are two types of knowledge captured in the ontology: (1) domain concepts, in which the ontology encodes concepts about risk factors, including ACEs (eg, abuse), SDoH (eg, housing condition), health outcomes such as chronic diseases (eg, asthma) and stress, and interventions; and (2) semantic inference, in which some of the knowledge in the ontology is encoded in the form of axioms. The axioms express knowledge related to (1) inclusions that define how two concepts are related; (2) equivalence relationships; and (3) causal knowledge, which are statements in the form of assertions of links between risk factors and health outcomes (eg, "obesity is a risk factor of diabetes" and "stress and exposure to toxins are risk factors for asthma").

### Knowledge-Based UPHO

The UPHO is a knowledge-based repository that can be used to represent and infer neighborhood-level indicators (eg, blight prevalence, poverty, education, proximity to clinics, proximity to public transportation) that may lead to negative health outcomes (eg, asthma, diabetes, stress, obesity). Moreover, the UPHO provides an analytics layer for calculating several metrics (eg, affinity scores in chronic conditions as a measure for comorbidity [16]) from analyzing neighborhood data.

### Objectives

We aimed to develop a knowledge-driven evidence-based recommendation system and a digital assistant to facilitate mental health surveillance. We first present our methodology along with the general architecture of the proposed recommendation system. We then demonstrate the feasibility and usability of our approach through multiple use case scenarios and offer recommendations for further development.

## Methods

### Platform Design

The idea behind the SPACES platform is to monitor the causes of ACEs and SDoH, and their impacts on health. This platform is based on the ACESO to provide the contextual knowledge needed to facilitate intelligent exploratory and explanatory analysis. Through this framework, decision makers can (1) identify risk factors, (2) integrate and validate ACEs and SDoH exposure at individual and population levels, (3) and detect high-risk groups. The idea for implementing the recommendation system for surveillance of ACEs was motivated by a study conducted under the Family Resilience Initiative (FRI) at Le Bonheur Children's Hospital (Memphis, Tennessee) that serves families with children during regular child visits in the clinic. Our use case scenarios to demonstrate the main components and features that constitute the SPACES framework were inspired by client examples and typical issues (Textbox 1), and the follow-up activities (Textbox 2) reported in the FRI reports.

**Textbox 1.** Typical adverse childhood experiences (ACEs)- and social determinants of health (SDoH)-related risk factors that arise in Family Resilience Initiative reports.

ACEs
- Child behavioral issues
- Child developmental health

SDoH
- Housing
- Food insecurity
- Transportation
- Education
- Legal/benefits

**Textbox 2.** Typical follow-up activities in Family Resilience Initiative reports.

Well-being check-in

Following up on a referral

Renewal inquiry

Client assistance

Contact resources on behalf of a client

Sharing information about future training

Confirming appointment (psychological, clinical, education, legal)

Arranging transportation

Scheduling appointment (psychological, clinical, education, legal)

## System Architecture

The main components of the SPACES framework are illustrated in Figure 2. A detailed explanation of each component is provided below.

**Figure 2.** Architecture of the Semantic Platform for Adverse Childhood Experiences (SPACEs) that reflects the main components of the system. The blue components are the services that we implement for the mental health domain, and the green components reflect general components that can be generalized to any other domain. QA: question-answering; ACEs: adverse childhood experiences; SDoH: social determinants of health.

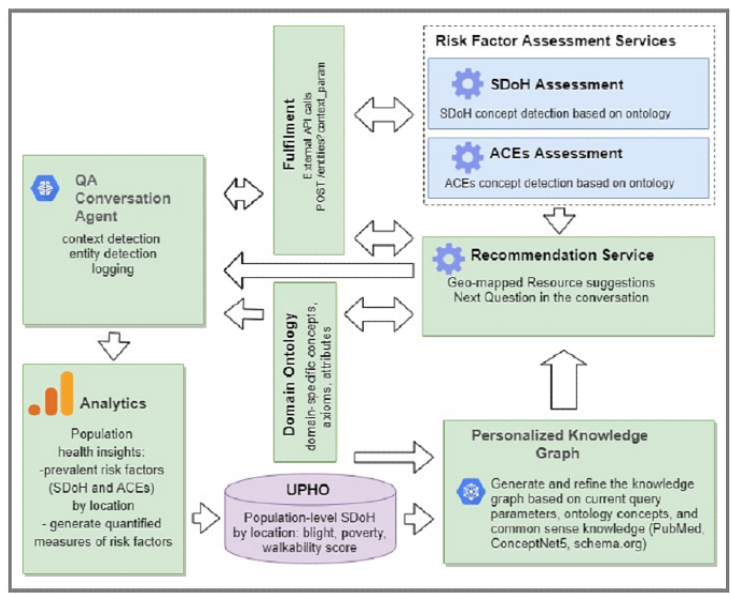

### QA Conversation Agent

To implement the conversation agent, we used the Google DialogFlow framework [17] and defined the following constructs.

*Intents*

An intent represents the purpose of a user's input. We start by defining a set of intents and supplying those with training phrases. This trains the conversation agent on detecting an intent based on values (eg, mold) that represent entity types (eg, @housing_circumstance) tagged in the text. For our task, we define a generic FRI_Assessment intent that is detected when the user enters a text similar to a training phrase. This intent has the following child intents corresponding to different risk factors (eg, housing, food, transportation) and follows up associated activities.

The SDoH_surveillance intent is detected whenever the text has phrases related to entity types that match SDoH-related concepts in the ontology. For instance, the entity type housing_circumstances is detected whenever the entity values "mold," "lead-based paint," "inadequate heating," and others are tagged in the text. Since all of these concepts are SDoH-related risk factors, the SDoH_Surveillance follow-up intent is also detected.

The ACEs_surveillance intent is detected whenever the text includes phrases related to entity types that match ACEs-related concepts in the ontology.

The FRI_followup_activity intent is triggered whenever an action is detected in the text (eg, schedule an appointment).

*Entity Types*

Entity types are ontological concepts that dictate how data are extracted from the user's raw text. For instance, the entity type housing_circumstance is detected whenever the entity values "mold," "lead-based paint," "inadequate heating," and others are tagged in the text. We load entity types from the ontology into the agent, and then enhance the agent with a minimal set of training phrases to enable it to tag entity types that appear in the phrases. Figure 3 shows a sample training phrase tagged with entity types.

**Figure 3.** A sample training phrase and detected contextual parameters with entity types and values.

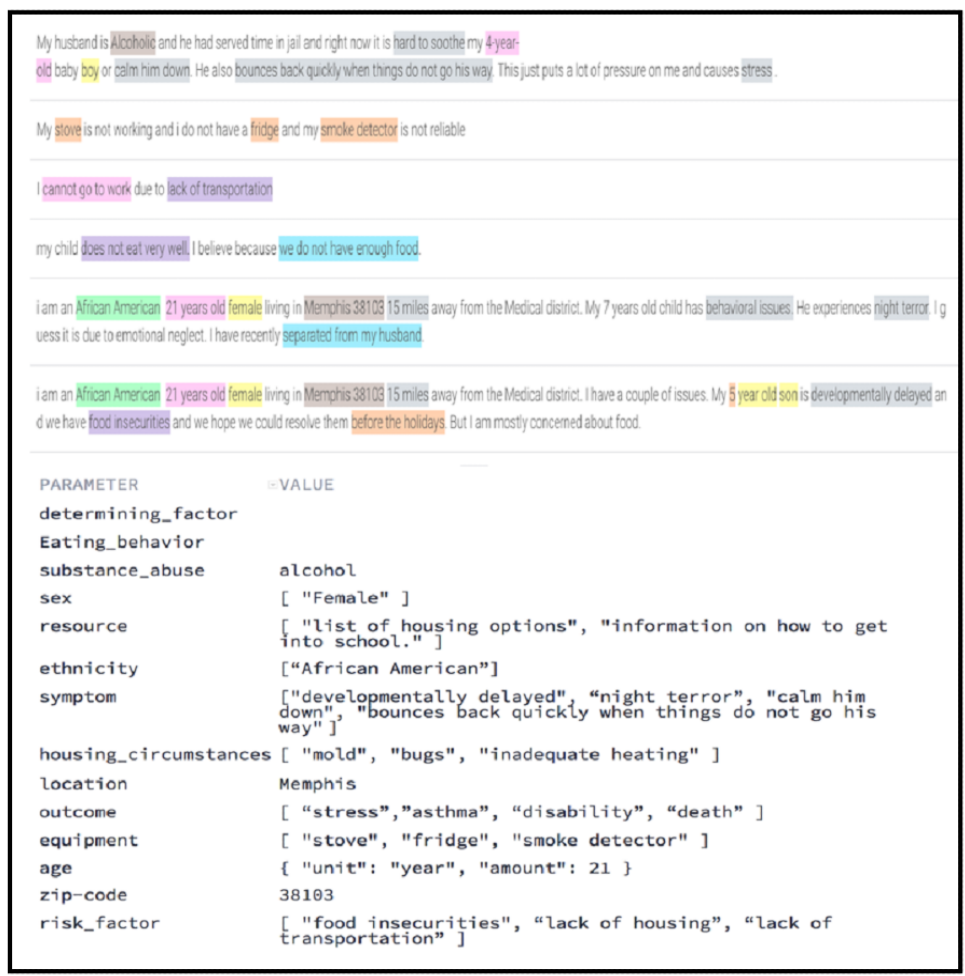

*Contextual Parameters*

Parameters are structured data extracted from raw text and have types that correspond to the entity types defined in the ontology. When an intent is matched at runtime, the agent extracts those parameters from entity type values (eg, 38103=zip code; night terror=symptom) that appear in the user expression (Figure 3). The agent then uses those parameters to perform logic and generate responses, or they can be exchanged between different contexts to control the conversation flow.

### Contexts

To keep track of the conversation flow, the agent maintains all active contexts (see connection arrows in Figure 4) on a stack to make sure they remain active throughout the conversation. At each point in the conversation, either a new intent or a follow-up child intent is detected due to a user's input or an event and fulfillment configurations within those intents. There are different types of contexts: (a) detected intent contexts, which are triggered by training phrases; (b) follow-up intent contexts, which are triggered by their parent intent or whenever the parent intent is triggered for fulfillment; (c) slot-filling context, which becomes active when the user does not provide values for preconfigured mandatory parameters; and (d) input/output contexts, in which intents can be configured with input and output contexts. A parent intent is by default an input context for its child intent and a follow-up intent is an output context for its parent.

**Figure 4.** Sample conversation flow to demonstrate the different constructs used to define the question-answering (QA) agent, including intents, contexts, events, fulfillments, and webhooks. ACEs: adverse childhood experiences; UPHO: Urban Population Health Observatory; FRI: Family Resilience Initiative; SDoH: social determinants of health.

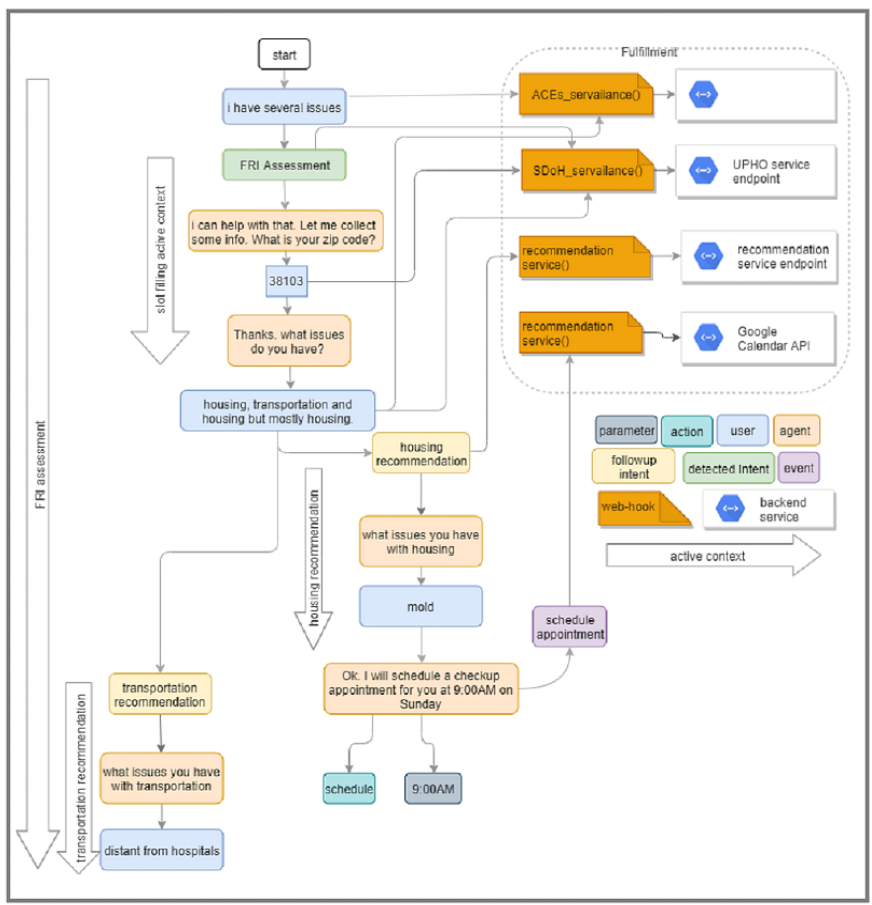

### Fulfillments and Webhooks

Fulfillments and webhooks enable the agent to invoke external service endpoints and send dynamic responses based on user expressions as opposed to hard-coding those responses. Fulfillment for an intent can be enabled by setting up a webhook, which is a service endpoint that we create and host. The agent sends a webhook request message that contains information about the matched intent, action, parameters, and response defined for the intent to one of our webhook services. The webhook service performs actions as needed (eg, query the knowledge graph or invoke external application programming interfaces [APIs]). The service then sends a webhook response message to the agent, which sends it to the end user.

### Events

An intent could be detected either by a phrase in a user's text or by being configured for an event. Fulfillments can be used to invoke external APIs. When the agent receives a webhook response (from a backend API) that includes an event, it immediately triggers the intent in which that event is defined.

Figure 5 illustrates a sample conversation flow that shows how intent detection and context activation occur, and how the different entities within the QA agent communicate

**Figure 5.** Two scenarios for conversation flow. (a) Scenario 1: client does not provide much detail and is prompted with mandatory parameters asked in a certain order. (b) Scenario 2: The client provides details for mandatory parameters. Since the user's text contains lead-based paint, the agent makes a hypothesis that a household member might need an early diagnosis for asthma based on the fact that lead-based paint is a toxicant and that exposure to toxicants may lead to asthma. The parameter values get substituted at run time and more intents get detected while active contexts remain and new contexts get added. To keep track of where the user is in the conversation, the agent keeps track of all active contexts on a stack.

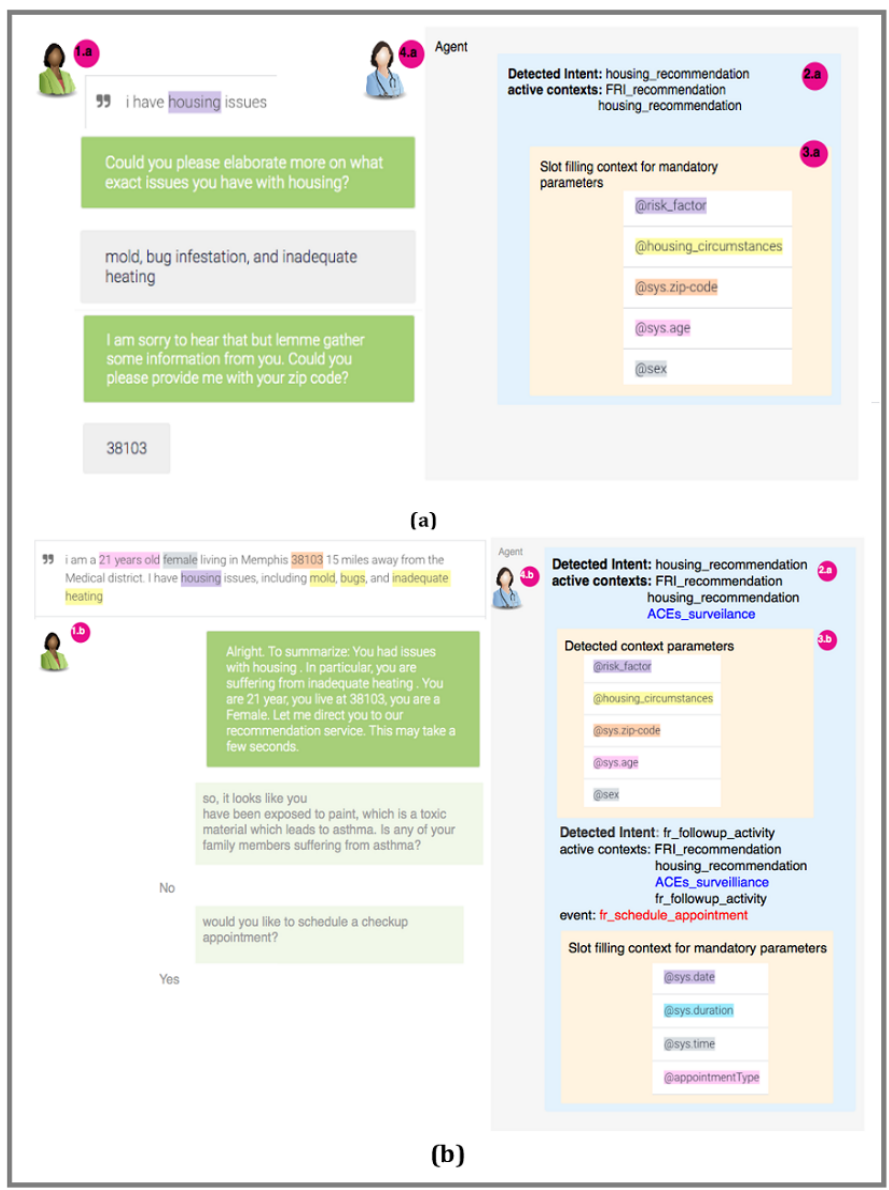

The user types or says an expression, which might be either detailed (eg, "I am an African American female and I have housing issues") or vague (eg, "I have several issues"). The presence or lack of details triggers different contexts. In either case, the agent matches the user expression to the generic FRI_Assessment intent, which is configured with a fulfillment that enables the agent to send a webhook request to one of the webhook services (ie, ACEs surveillance, SDoH surveillance, recommendation service). In the case of vague user input, the agent detects a slot-filling context by prompting the user with extra questions until they have provided values for all required parameters. In the case of a detailed user input, the agent directly moves to a follow-up intent, which might as well be configured with a fulfillment. After filling values for contextual parameters, the agent sends a webhook request to the recommendation service. The service responds with a webhook response that includes either a resource or a follow-up question based on the knowledge graph. A phrase provided by a user may contain parameters (eg, zip code), actions (eg, schedule an appointment), or priority words (eg, "I am interested in furthering my education, but would prefer a job first"). If a user's input includes more than one issue, then both follow-up intents are detected, but the agent handles them one at a time either based on the order they were mentioned or by using priority words.

Since the FRI_Assessment is configured with a fulfillment, the agent proceeds as though the end user initiated the match for the FRI_followup_activity intent. Thus, instead of responding to the user for the FRI_Assessment intent match, the agent triggers the FRI_followup_activity intent, which is configured for the event "schedule an appointment." Finally, the FRI_followup_activity intent handles the required parameters (date, duration, time, and appointment type) and fulfillment (eg, Google calendar API) as dictated by the configuration of FRI_followup_activity intent.

### SDoH Surveillance Service

The SDoH surveillance service is triggered whenever an SDoH-related entity type is introduced in the user's conversation with the QA agent. To achieve this, the surveillance uses the ACESO to infer whether a detected entity type (eg, being_exposed_to_lead-based_paint) is a subtype of the more generic SDoH type as well as the causal knowledge (eg, lead-based_paint *causes* asthma). This service also invokes the UPHO service endpoint to obtain metrics (eg, blight prevalence, walkability score) based on the current user's neighborhood.

### ACEs Surveillance Service

This service keeps track of all possible questions that can be asked during the ACEs assessment process [18]. It is triggered each time an ACE-related entity type is introduced in the user's conversation with the agent. It gets invoked by the recommendation service to retrieve only the questions relating to the ACEs concepts provided by the agent. The QA agent keeps track of how many ACEs questions were reported as positive, and provides those to the ACEs surveillance service, which uses the knowledge stored in the ACESO about ACEs score classification and rules for calculating them. It then sends the resulting score to the recommendation service, which seeks insights from external knowledge (eg, research publications [18]) to formulate a diagnosis based on the question: "given an ACEs score of X and symptoms S1,…Sn, what are the likely negative outcomes to screen for?"

### Recommendation Service

To make real-time recommendations, this service instantly captures new resource interests, ACEs, and SDoH-associated risk factors detected in the user's current conversation and uses them to incrementally refine a personalized knowledge graph. At each stage in the conversation, the QA agent passes detected entity types and contextual parameter values to the recommendation service. Entity types help the service determine entry points on the knowledge graph, and contextual parameters help refine the queries further to obtain a more personalized version of the graph (Figure 6). Once the personalized graph is generated, the service supplies the QA agent with two types of recommendations: the next question to ask and a resource to suggest.

**Figure 6.** A personalized graph is generated based on contextual parameters provided by Alice (see Scenario 4 in Textbox 3) after populating ontology concepts with real-time contextual parameters supplied by the agent, and after enriching the graph with external common sense knowledge. The figure illustrates how a concrete path on the graph leads to a recommendation and the metapath that can be derived from that path.

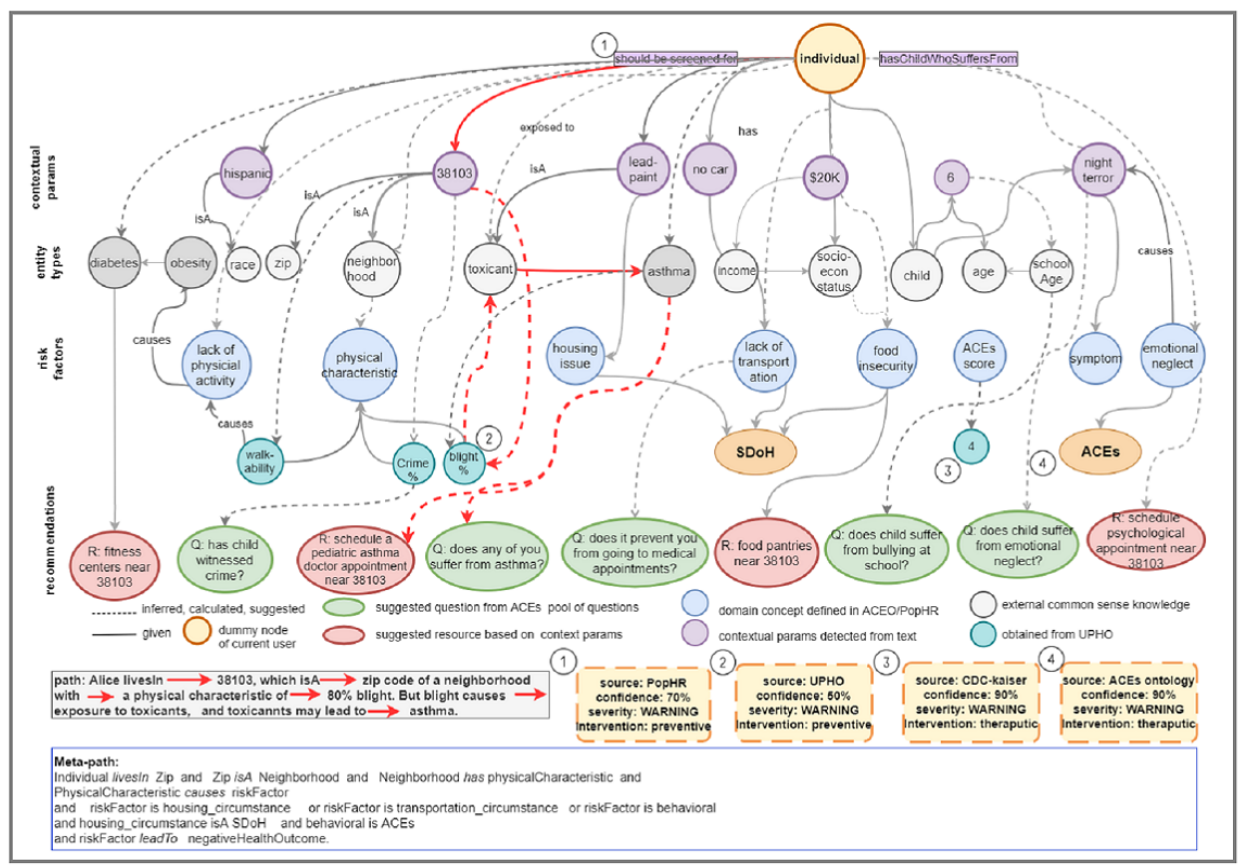

We used Neo4j graph technology [19] for generating and enriching the graph. Before populating the graph with real-time data, we loaded ACEs, SDoH, and health outcome concepts from the ontology into a graph and persist them. We used the NeoSemantics framework [19] to import the Resource Description Framework (RDF)-based data model schemas of the ontology as a metagraph into a Neo4j graph. The resulting property graph inherits the modeling limitations of the RDF, including the lack of support for attributes on relationships. Therefore, we enriched and fixed the raw graph after loading it in Neo4j. To populate the graph, the service starts with a *dummy node* that represents the current user (Figure 6) and then incrementally adds the following node types.

Domain concept nodes correspond to *entity types* detected by the QA agent (eg, race, age, zip code, symptom). Entity type nodes could be either (a) *risk factors* such as "housing_circumstance," "food_issue," and "household_issue," or (b) common sense knowledge such as school age and domain concepts in the given context (eg, age *is a* school age and lead-based paint *is a* toxicant).

Value nodes are populated with real-time values either (a) from *contextual parameters* obtained from the QA agent (eg, "38103," "night terror"), (b) calculated based on the ontological axioms (eg, ACEs score of 4), or (c) retrieved from the UPHO (eg, 80% blight prevalence).

Recommendation nodes are *question* nodes and *resource* nodes. Question nodes represent follow-up questions suggested for the QA agent. The answers to these questions coming from the QA agent can further refine the graph. The suggested question nodes are solicited from a pool of questions kept by the ACEs surveillance service. Resource nodes (eg, referrals, housing options, school information, doctors, clinics) are resources suggested for the QA agent to provide to the user. These resources are suggested based on the user's demographic profile (eg, the food banks within a small radius from their zip code). Resource nodes are pulled from a pool of existing resources or by invoking external backend APIs. In addition to the nodes pulled from contextual parameters, the graph shows nodes that are added based on knowledge inferred through axioms (dashed arrows in Figure 6) and concept hierarchies (solid arrows in Figure 6).

**Population Health and Policymaking Analytics**

The analytics component utilizes the conversation history recorded by the DialogFlow logging API, including (1) timestamps, number of interactions (within a conversation) for all user sessions, and percentage of mismatches if any; (2) a visual summary of the conversation flow with percentages for each detected intent as well as the conversational paths that users have taken when interacting with an agent; (3) popularity per intent by showing the number of sessions in which the intent was matched as well as the number of times the intent was used (total from all sessions); (4) percentage of sessions in which a user exited the conversation in the specified intent compared to the total number of sessions in which this same intent was matched; and (5) average response time to user requests.

The aggregated results from all user conversations provide policymakers with insights about population health. Analyzing such data can help in designing interventions and preventive measures based on the most prevalent risk factors in certain regions. It can also assist the framework users by providing recommendations on how to direct future conversations. For example, it can look into smaller communities within geographic areas or perform collaborative filtering based on similarities in user behavior, or detect similar communities on the conversation graph. For example, if a user that belongs to a certain age or ethnicity group shows a certain pattern/route during the case assessment conversation procedure, then the QA agent may suggest the same route for the next user with similar criteria.

## Results

We describe the main features provided by SPACES through a proof-of-concept prototype that will render the information collected by the QA agent and the recommendation service on a user-friendly interface. The prototype is intended for several types of users, including caregivers (eg, child-parents) or health care professionals (eg, nurses, physicians, social workers). The main features of the view that would be available to a health care professional are illustrated in Figure 7, including recommendations for digital assistance (A), studying the association between ACEs and SDoH (B.2 and B.5), knowledge graph querying (B.3), geocoded resource recommendation (B.6), and explainability by displaying inference sources (B.5 and B.7). The QA agent view that would be available to a caregiver is illustrated in Figure 5 and in panel A of Figure 7.

We present multiple use-case scenarios in Textbox 3. For simplicity, we use Scenario 4 to demonstrate how the QA agent detects context and lays it over to the recommendation service, and how the recommendation service uses contextual parameters to generate a personalized knowledge graph.

**Figure 7.** Prototype of the recommendation system. (A) Question-answering (QA) view (1) send/revise suggested questions. (B) Health care practitioner assessment panel, including (2) social determinants of health (SDoH) detection, (3) knowledge graph and queries, (4) pool of previously asked questions,(5) alerts for detected adverse childhood experiences (ACEs) symptoms, (6) geocoded resource allocation, (7) explain recommendations, (8) and visual analytics.

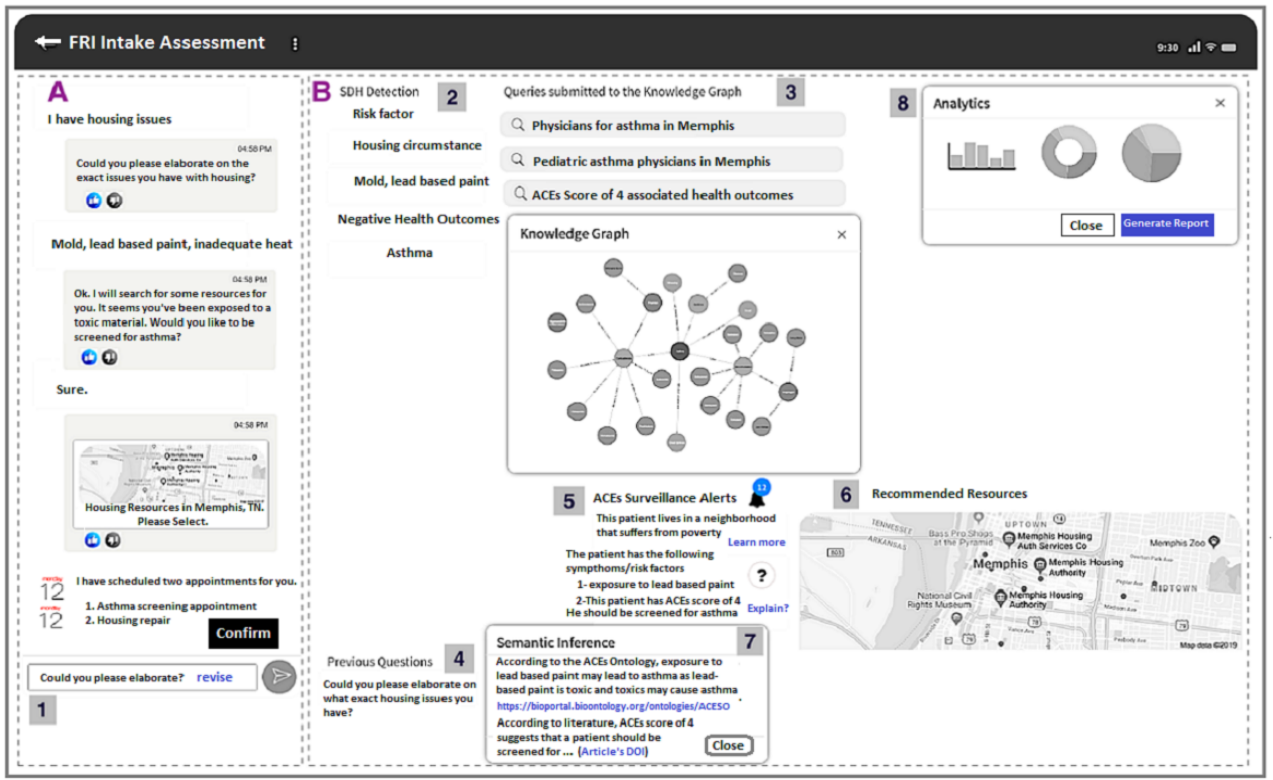

**Textbox 3.** Four use case scenarios.

---

**Scenario 1: Assessing needs relating to SDoH. SDoH: social determinants of health; N/A: not applicable; ACEs: adverse childhood experiences; UPHO: Urban Population Health Observatory.**

"I am currently residing in a safe place, but I'm concerned about my household income as I am currently unemployed due to legal issues. I have some college and I am interested in furthering my education, but would prefer a job first."

*Symptoms:* N/A

*Risk factors:* ACEs detected: Unemployed due to legal issues (Legal, Textbox 1)

    I'm interested in furthering my education (Education, Textbox 1)

*Outcomes:* N/A

*Intervention*: Information on how to get into the school

    Follow up on legal issues.

**Scenario 2: Assessing issues relating to ACEs**

"My husband is an alcoholic and he had served time in jail and right now it is hard to soothe my 4-year-old baby boy or calm him down. He also bounces back quickly when things do not go his way. This just puts a lot of pressure on me"

*Symptoms:* Hard to soothe or calm down (behavioral)

    Bounces back quickly (behavioral)

*Risk factors:* ACEs detected: living in a household with substance abuse

    ACEs detected: living with a household member who was in jail

*Outcomes:* provided (stress)

    inferred (asthma)

*Intervention:* Therapeutic (schedule psychologist appointment)

**Scenario 3: Mixed model of ACEs and SDoH**

"I have a couple of issues. My 7-year-old son is developmentally delayed, and we have food insecurities that we hope we could resolve before the holidays. But I am mostly concerned about food."

*Symptoms*: Child developmental delay (behavioral)

*Risk factors:* food insecurities

*Outcomes:* N/A

*Intervention:* Provide information about food pantries

    Schedule psychologist appointment

**Scenario 4: Detecting risk factors for potential negative health outcomes and providing early diagnosis**

"I am a Hispanic 21-year-old female living in Memphis. My 6-year-old child experiences night terror. I have recently separated from my husband."

*Symptoms:* Night terror (emotional neglect causes night terror)

*Risk factors:* ACEs detected:

    provided (living in a household of divorce)

    inferred (ontology, emotional neglect)

    (UPHO, a neighborhood with blight)

*Outcomes:* inferred (asthma)

*Intervention:* Therapeutic (schedule a medical appointment)

---

Based on Scenario 4, Alice is a Hispanic female located in a city with zip code 38103 and has a 6-year-old child who suffers from night terrors. Alice can either provide vague (Figure 5a) or detailed (Figure 5b) text. Either way, the agent collects as many contextual parameters as possible and then passes them over to the recommendation service.

The service starts building the personalized graph for Alice by adding a node labeled *zip code* and *has_value* 38103. It then links the node to all concepts related to a zip code, including neighborhood and physical characteristics. The physical characteristics are then linked to all related concepts based on the ACESO, including blight, walkability, and crime, and it populates these nodes with values from the UPHO. If blight has the maximum value, it links the blight concept with related concepts based on the axiom (blight causes exposure to toxicants and exposure to toxicants leads to asthma). Thus, it links it to the toxicant and asthma nodes. Additionally, it adds the age node and populates it with a contextual parameter value of 6, and links the age concept to the school-age concept. It then adds the *symptom* node and links it to *night terror* based on the concept subtype relation in the ontology. It then links *night terror* to *emotional neglect* based on an axiom. The inference path in Alice's scenario is shown on the graph in red in Figure 6 to illustrate how Alice's case leads to a recommendation for pediatric asthma physicians. The metapath, derived from that concrete path, can be used in future recommendations if an individual is encountered with similar contextual parameters.

The resulting graph is a property graph in that both nodes and relations have properties. For instance, the inferred relations (eg, *shouldBeScreenedFor* asthma, has an ACEs score of 4, and *livesIn* a neighborhood with 80% blight prevalence) can be labeled with the source of inference, confidence probability, severity, and type of intervention. Sources of knowledge inference appear in yellow boxes at the bottom of Figure 6, which include UPHO, research papers, the ACESO, and others. We can enrich the graph further with a clinic locator graph of physicians who can provide prescriptions for pediatric asthma, and then filter the resource nodes further based on provided contextual parameters. For instance, we can keep only clinics that fall within a small radius from Alice's residential zip code and that provide Spanish language–speaking services based on her ethnicity.

The health care practitioner can observe the inferred knowledge through several features rendered on the prototype interface as follows: (1) ACEs alerts (eg, Alice's case indicates an ACEs score of 4, night terror as a symptom, and suggests an early diagnosis of asthma as an outcome) (Figure 7, B.2); (2) SDoH alerts (eg, the risk factors associated with Alice's case) (Figure 7, B.5); (3) visual aids for explainability by clicking on the (?) icons to display links to inference sources (eg, the ontology link on BioPortal and the identifier for the CDC-Kaiser paper) (Figure 7, B.7); (4) geocoded view of the suggested resources (eg, clinics in Memphis) (Figure 7, B.6); and (5) which questions to ask next (Figure 7, B.4).

## Discussion

### Principal Findings

The significance of the proposed approach lies in its ability to provide recommendations to the QA agent with the least effort from both the user and the health care practitioner. It aims to maximize knowledge about the patient without having to delve into all of the questions that are often asked in ACEs and SDoH intake assessments. It also provides the ability to explain why a certain question or resource was suggested.

Preliminary rapid prototyping of the recommender system allowed for early verification of the functionalities through the multiple case scenarios described in this paper. We anticipate rapid feedback from end users on various features during an iterative development process, and finally establish a comprehensive usability and user experience test. We have evaluated the SPACES semantic framework and its underlying ontology automatically using description logic reasoners such as Fact++ [20] to ensure the consistency and satisfiability of the ontology and semantic model. We also used inputs from our collaborating domain experts to assess the soundness and completeness of the ontology by examining how well it has been aligned with the required criteria defined in our domain and scope. Furthermore, we evaluated the usability of ACESO based on its functionality and capability to respond and answer the target queries. Finally, further work is underway to develop and conduct a series of formal evaluation practices for a comprehensive assessment of the accuracy, usability, coverage, confidence, trustability, as well as adaptivity and scalability of the recommender system [21]. Moreover, we will assess the utility and impact of the system on the surveillance of ACEs to generate a timely response and intervention, ultimately informing public health planning and policymaking.

The proposed approach might face some limitations. One limitation is in providing an overall guiding architecture to support transfer ability between health domains. Several of the QA platforms (eg, Google DialogFlow and IBM Watson) read rules on how to answer questions from backend sources (eg, HTML FAQ files, Plaintext files). These sources can help load the questions into the QA agent. They also require training the agent with concepts that may appear in the QA text. The ontology in our case helps load the concepts automatically, which is separate from the QA platform implementation itself. Each domain will have its own concepts that can be encoded in a separate ontology, and we can either develop new ontologies or reuse existing ones.

Another limitation is that the recommendation system has access to only population-wide data, where the population's characteristics might be different than the characteristics of the individuals living in that population or neighborhood. However, for specific users and specific use cases, it needs access to individual data. For instance, a pediatrician trying to decide whether a child is suffering from ACEs will need to have access to the child's health history and other relevant information but should not be allowed to access information about the parent's finances or criminal history (outside of what is publicly available). Moreover, a local judge will want to have access to the criminal history of the family members if they want to decide whether the child should be removed from their parents to ensure their safety, but they should not be able to access any of their medical records. Thinking about how to control mediated access to sensitive information will be a key part of the development of the recommendation system. We are currently working on integrating the recommendation system into a personalized health library [22].

The adoption of recommendation systems may be hindered by a poor user-interface design or poor integration into clinical workflows. Human factors engineering can improve efficiency, reduce errors, increase technology adoption, and reduce the early abandonment of systems. This paper lacks an objective evaluation of *how* health care practitioners will benefit from the explanations or how the quality of those explanations would be assessed. We plan to use an iterative user-centered design and formative evaluation by conducting predevelopment focus groups, which might reveal issues related to manual data entry from target users and the time spent reviewing generated knowledge from providers. Ongoing research on this project will also involve implementing an optimization algorithm of the recommendations. We will also consider technically evaluating the system for precision and performance.

All updates in the underlying ontologies and semantic structure will be managed through our previously implemented framework [23, 24]. As for the target audience, the system is intended for a variety of users depending on the domain, including social workers, health care providers (eg, nurses, physicians), and caregivers (child-parent). The digital assistant or QA agent is the part intended and available for end users (eg, caregivers). It is a simulation of the conversations that occur between health care providers or social workers and caregivers as they discuss their case. The text exchanged through the chat is used in the backend for both enhancing the QA agent and refining the personal knowledge graph of the current user. The visual graphs and underlying reasoning are intended for AI explainability, which is critical for health care providers to make decisions. In particular, the analytics part is intended for monitoring and research purposes, and therefore is more appropriate for health professionals and policymakers.

Finally, we discuss the implications for policy and practice. We believe that early intervention is the best way to prevent the progression of negative health outcomes to their end stage, and that well-designed early detection systems can aid clinicians by generating knowledge that can be aligned with clinical workflows. Thus, systems tailored toward such interventions by utilizing knowledge about self-reported or detected SDoH and ACEs would be most useful. Through the framework, decision makers can (1) identify risk factors, (2) integrate and validate ACEs and SDoH exposure at individual and population levels, and (3) detect high-risk groups. The analytics component could be most useful for policymakers for this purpose and is intended for monitoring and research purposes.

## Conclusions

In this study, we leveraged explainable AI to present a proof-of-concept prototype for a knowledge-driven evidence-based recommendation system to improve the surveillance of ACEs. The proposed approach will enhance the health care practitioner's ability to provide explanations for the decisions that they make. Further development and official evaluation are underway to include a privacy layer through a personal health library and to conduct a clinical trial for formally assessing both the usability and usefulness of the implementation.

### Acknowledgments

We would like to thank Dr Robert L Davis, Dr Jonathan A McCullers, Dr Jason Yaun, Dr Sandra R Arnold, and the entire team at the Family Resilience Initiative at Le Bonheur Children's Hospital, Memphis, Tennessee, for their support and insights. This research was partially supported by the Memphis Research Consortium.

### Conflicts of Interest

None declared.

### References

1. Felitti VJ, Anda RF, Nordenberg D, Williamson DF, Spitz AM, Edwards V, et al. Relationship of childhood abuse and household dysfunction to many of the leading causes of death in adults. The Adverse Childhood Experiences (ACE) Study. Am J Prev Med 1998 May;14(4):245-258. [doi: 10.1016/s0749-3797(98)00017-8] [Medline: 9635069]
2. Social Determinants of Health. World Health Organization. 2019. URL: https://www.who.int/social_determinants/sdh_definition/en/ [accessed 2020-09-17]
3. Shin EK, Shaban-Nejad A. Urban Decay and Pediatric Asthma Prevalence in Memphis, Tennessee: Urban Data Integration for Efficient Population Health Surveillance. IEEE Access 2018;6:46281-46289. [doi: 10.1109/access.2018.2866069]
4. Chung EK, Siegel BS, Garg A, Conroy K, Gross RS, Long DA, et al. Screening for Social Determinants of Health Among Children and Families Living in Poverty: A Guide for Clinicians. Curr Probl Pediatr Adolesc Health Care 2016 May;46(5):135-153 [FREE Full text] [doi: 10.1016/j.cppeds.2016.02.004] [Medline: 27101890]
5. Shorey S, Ang E, Yap J, Ng ED, Lau ST, Chui CK. A Virtual Counseling Application Using Artificial Intelligence for Communication Skills Training in Nursing Education: Development Study. J Med Internet Res 2019 Oct 29;21(10):e14658 [FREE Full text] [doi: 10.2196/14658] [Medline: 31663857]
6. Battaglia P, Hamrick J, Bapst V, Sanchez-Gonzalez A, Zambaldi V, Malinowski M, et al. Relational inductive biases, deep learning, and graph networks. arXiv preprint 2018 Oct 17 [FREE Full text]
7. Ehrlinger L, Wöß W. Towards a Definition of Knowledge Graphs. In: SEMANTiCS (Posters, Demos, SuCCESS). 2016 Presented at: 12th International Conference on Semantic Systems; September 12-15, 2016; Leipzig, Germany.
8. Ruan T, Huang Y, Liu X, Xia Y, Gao J. QAnalysis: a question-answer driven analytic tool on knowledge graphs for leveraging electronic medical records for clinical research. BMC Med Inform Decis Mak 2019 Apr 01;19(1):82 [FREE Full text] [doi: 10.1186/s12911-019-0798-8] [Medline: 30935389]
9. Goodwin TR, Harabagiu SM. Medical Question Answering for Clinical Decision Support. Proc ACM Int Conf Inf Knowl Manag 2016 Oct;2016:297-306 [FREE Full text] [doi: 10.1145/2983323.2983819] [Medline: 28758046]
10. Rotmensch M, Halpern Y, Tlimat A, Horng S, Sontag D. Learning a Health Knowledge Graph from Electronic Medical Records. Sci Rep 2017 Jul 20;7(1):5994. [doi: 10.1038/s41598-017-05778-z] [Medline: 28729710]

11. Nelson CA, Butte AJ, Baranzini SE. Integrating biomedical research and electronic health records to create knowledge-based biologically meaningful machine-readable embeddings. Nat Commun 2019 Jul 10;10(1):3045. [doi: 10.1038/s41467-019-11069-0] [Medline: 31292438]
12. Brenas JH, Shin EK, Shaban-Nejad A. Adverse Childhood Experiences Ontology for Mental Health Surveillance, Research, and Evaluation: Advanced Knowledge Representation and Semantic Web Techniques. JMIR Ment Health 2019 May 21;6(5):e13498 [FREE Full text] [doi: 10.2196/13498] [Medline: 31115344]
13. Brenas JH, Shin EK, Shaban-Nejad A. An Ontological Framework to Improve Surveillance of Adverse Childhood Experiences (ACEs). Stud Health Technol Inform 2019;258:31-35. [Medline: 30942708]
14. Brenas JH, Shin EK, Shaban-Nejad A. A Hybrid Recommender System to Guide Assessment and Surveillance of Adverse Childhood Experiences. Stud Health Technol Inform 2019 Jul 04;262:332-335. [doi: 10.3233/SHTI190086] [Medline: 31349335]
15. Adverse Childhood Experiences Ontology. BioPortal. 2019 Feb 25. URL: https://bioportal.bioontology.org/ontologies/ACESO [accessed 2020-09-17]
16. Shin EK, Kwon Y, Shaban-Nejad A. Geo-clustered chronic affinity: pathways from socio-economic disadvantages to health disparities. JAMIA Open 2019 Oct;2(3):317-322 [FREE Full text] [doi: 10.1093/jamiaopen/ooz029] [Medline: 31984364]
17. Dialogflow. Google Cloud. 2010. URL: https://dialogflow.com/ [accessed 2022-10-20]
18. Chang X, Jiang X, Mkandarwire T, Shen M. Associations between adverse childhood experiences and health outcomes in adults aged 18-59 years. PLoS One 2019;14(2):e0211850 [FREE Full text] [doi: 10.1371/journal.pone.0211850] [Medline: 30730980]
19. Neo4j Platform. URL: https://neo4j.com/ [accessed 2020-09-17]
20. Tsarkov D, Horrocks I. FaCT++ Description Logic Reasoner: System Description. In: Furbach U, Shankar N, editors. Automated Reasoning. IJCAR 2006. Lecture Notes in Computer Science, vol 4130. 2006 Presented at: International Joint Conference on Automated Reasoning; August 17-20, 2006; Seattle, WA URL: https://link.springer.com/chapter/10.1007%2F11814771_26 [doi: 10.1007/11814771_26]
21. Shani G, Gunawardana A. Evaluating Recommendation Systems. In: Ricci F, Rokach L, Shapira B, Kantor P, editors. Recommender Systems Handbook. Boston: Springer; 2011:257-297.
22. Ammar N, Bailey JE, Davis RL, Shaban-Nejad A. The Personal Health Library: A Single Point of Secure Access to Patient Digital Health Information. Stud Health Technol Inform 2020 Jun 16;270:448-452. [doi: 10.3233/SHTI200200] [Medline: 32570424]
23. Shaban-Nejad A, Haarslev V. Managing changes in distributed biomedical ontologies using hierarchical distributed graph transformation. Int J Data Min Bioinform 2015;11(1):53-83. [doi: 10.1504/ijdmb.2015.066334] [Medline: 26255376]
24. Shaban-Nejad A, Ormandjieva O, Kassab M, Haarslev V. Managing Requirement Volatility in an Ontology-Driven Clinical LIMS Using Category Theory. Int J Telemed Appl 2009;2009:917826. [doi: 10.1155/2009/917826] [Medline: 19343191]

## Abbreviations

**ACEs:** adverse childhood experiences
**ACESO:** Adverse Childhood Experiences Ontology
**AI:** artificial intelligence
**API:** application programming interface
**FRI:** Family Resilience Initiative
**ML:** machine learning
**QA:** question-answering
**RDF:** Resource Description Framework
**SDoH:** social determinants of health
**SPACES:** Semantic Platform for Adverse Childhood Experiences Surveillance
**UPHO:** Urban Population Health Observatory